\documentclass{article}
\pdfoutput=1
\usepackage{spconf,amsmath,graphicx,balance}
\usepackage{textcomp}
\usepackage{fancyhdr}
\title{Spatially adaptive image compression using a tiled deep network}
\name{D. Minnen, G. Toderici, M. Covell, T. Chinen, N. Johnston, J. Shor,
  S.J. Hwang, D. Vincent, S. Singh}

\address{Google Inc., 1600 Amphiteatre Pkwy., Mountain View, CA 94043, USA}

\begin{document}
%
\maketitle
\begin{abstract}
  Deep neural networks represent a powerful class of function approximators that
can learn to compress and reconstruct images. Existing image compression
algorithms based on neural networks learn quantized representations with a
constant spatial bit rate across each image. While entropy coding introduces
some spatial variation, traditional codecs have benefited significantly by
explicitly adapting the bit rate based on local image complexity and visual
saliency. This paper introduces an algorithm that combines deep neural
networks with quality-sensitive bit rate adaptation using a tiled network. We
demonstrate the importance of spatial context prediction and show improved
quantitative (PSNR) and qualitative (subjective rater assessment) results
compared to a non-adaptive baseline and a recently published image compression
model based on fully-convolutional neural networks.

\end{abstract}

\fancypagestyle{firststyle}
{
   \fancyhf{}
   \lfoot{978-1-5090-2175-8/17 \textcopyright 2017 IEEE}
   \setcounter{page}{2796}
   \cfoot{\thepage}
   \rfoot{ICIP 2017}
}

\thispagestyle{firststyle}

\fancyhf{}
\renewcommand{\headrulewidth}{0pt}
\setcounter{page}{2796}
\cfoot{\thepage}

\section{Introduction}
\label{sec:intro}

Many researchers have investigated the use of neural networks to learn models
for lossy image compression (see~\cite{Jiang1999} for a review) including a
recent resurgence due to improved methods for training deep
networks~\cite{Toderici2016, balle2016, toderici2015variable,
  gregor2016conceptual}. These learned models produce compressed
representations with a fixed bit rate across the image. Some spatial variation
may be introduced by lossless entropy coding, which is applied as a
post-process to compress the generated representation. This variation,
however, is tied to the frequency and predictability of the codes, not
directly to the complexity of the underlying visual information.

\begin{figure}
  \centering
  \includegraphics[width=0.325\columnwidth]{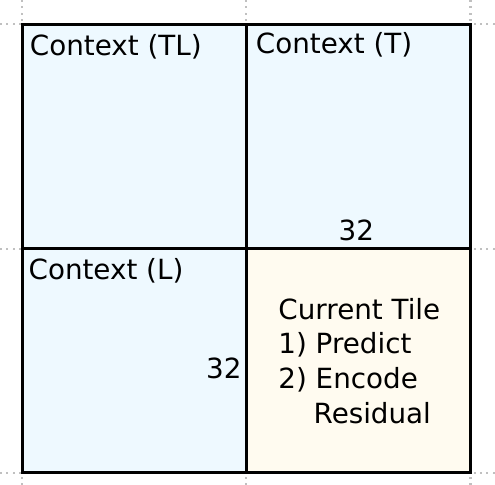}
  \hfill
  \includegraphics[width=0.325\columnwidth]{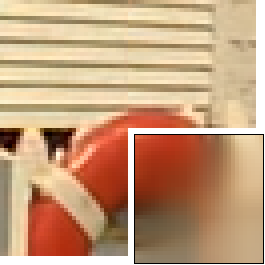}
  \hfill
  \includegraphics[width=0.33\columnwidth]{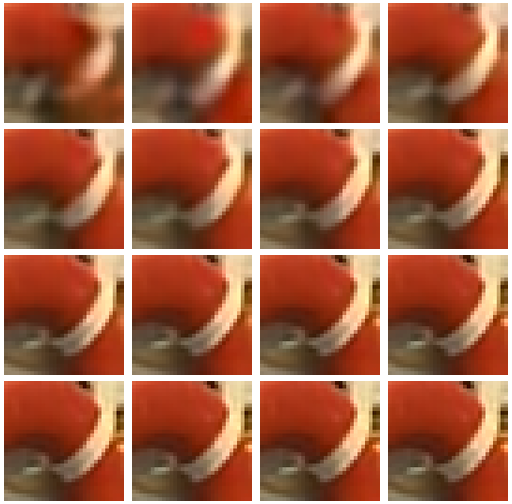}

  \caption{For each $32 \times 32$ tile, our model uses neighboring tiles
    above and to the left as context ({\it left}). First, a deep network
    predicts the pixel values for the target tile ({\it center}), and then a
    second network improves the reconstruction by progressively encoding the
    residual ({\it right}).}
  \label{fig:kermit-tiles}
\end{figure}

Traditional image codecs typically use both entropy coding and explicit bit
rate adaptation that depends on local reconstruction quality ({\it e.g.}, JPEG
2000, WebP, and BPG)~\cite{jpeg2000,WebP,BPG, bull2014}. This spatial
adaptation allows them to use additional bits more effectively by
preferentially describing regions of the image that are more complex or
visually salient.

This paper introduces an approach to image compression that combines the
advantages of deep networks with bit rate adaptation based on local
reconstruction quality. Neural networks provide two primary benefits for image
compression: (1) they represent an extremely powerful, nonlinear class of
regression functions ({\it e.g.}, from pixel values to quantized codes and
from codes back to pixels), and (2) their model parameters can be efficiently
trained on large data sets. The second benefit is particularly important
because it means that an effective architecture can be easily specialized to
new domains and specific applications. For example, an architecture that works
well on natural images can be retrained and optimized for cartoons, selfies,
sketches, or presentations, where each domain contains images with
substantially different statistics.

State of the art neural networks for image compression use fully-convolutional
architectures~\cite{Toderici2016, balle2016, toderici2015variable,
  gregor2016conceptual}. This design promotes efficient local information
sharing and allows the networks to run on images with arbitrary
resolution~\cite{Long2014}. The tradeoff is that the shared dependence on
nearby binary codes makes it difficult to adjust the bit rate across an image.
Research done in parallel to this paper investigates ways to overcome this
difficulty by using a more complex training procedure~\cite{covell2017}. Our
model, on the other hand, sidesteps the problem by using a block-based
architecture. This tiled design maintains resolution flexibility and local
information sharing while also significantly simplifying the implementation of
bit rate adaptation.


\section{Codec Overview}
\label{sec:overview}

Our method works by dividing images into tiles, using spatial context to make
an initial prediction of the pixel values within each tile, and then
progressively encoding the residual. This approach is similar to the
high-level structure of existing codecs such as WebP and BPG, though we use a
fixed $32 \times 32$ tiling while those methods use a more sophisticated
process for adaptively subdividing each image.

\begin{figure}
  \centering
  \includegraphics[width=\columnwidth]{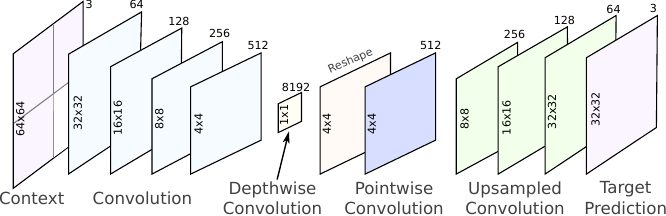}
  \caption{The context prediction network uses strided convolution to extract
    features from the context tiles and uses upsampled convolution to generate
    an RGB prediction for the target tile. Each block in the diagram
    represents a layer in the neural network with the resolution shown inside
    the block and the depth ({\it e.g.}, 3 for the RGB input and output) shown
    above.}
  \label{fig:context-predictor-arch}
\end{figure}

Image encoding proceeds tile-by-tile in raster order. For each tile, the
spatial context includes the neighboring tiles to the left and above (see
Figure~\ref{fig:kermit-tiles}). This leads to a $64 \times 64$ context patch
where the values of the target tile (the bottom-right quadrant) has not yet
been processed. The initial prediction for the target tile is produced by a
neural network trained to analyze context patches and minimize the $L_1$ error
between its prediction and the true target tile (details in
Section~\ref{subsec:context-prediction}). The goal is to take advantage of
correlations between relatively distant pixels and thus avoid the cost of
re-encoding visual information that is consistent from one tile to the next.

\begin{figure}
  \centering \includegraphics[width=0.9\columnwidth]{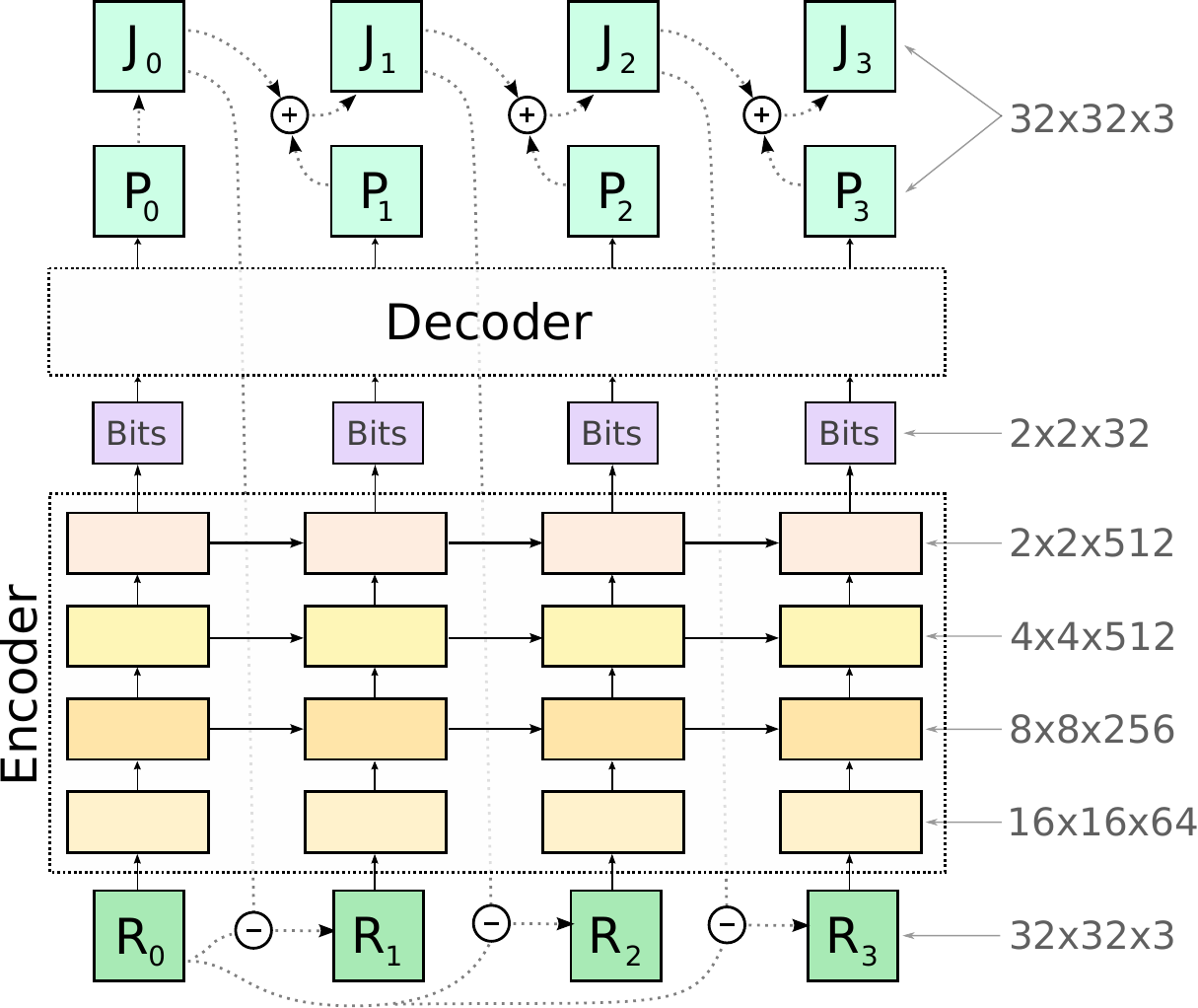}
  \caption{The residual encoder uses a recurrent auto-encoder architecture
    where each layer has the shape shown (height $\times$ width $\times$
    depth). Each iteration (four are shown) extracts features from its input
    ($R_i$) and quantizes them to generate 128 bits. The decoder learns to
    reconstruct the input from these binary codes. Each iteration tries to
    capture the residual remaining from the previous iteration so the sum
    across iteration outputs ($P_i$) provides a successively better
    approximation of the original input ($R_0 \approx J_i = \sum_{k=0}^i
    P_k$).}
  \label{fig:donald-rnn}
\end{figure}

Contextual data is unlikely to contain enough information to accurately
reconstruct image details or to predict pixel values across object
boundaries. The second step of our approach fills in such details by encoding
the residual between the true image tile and the initial prediction using a
deep network based on recurrent auto-encoders (details in
Section~\ref{subsec:residual-encoding}). After a tile has been encoded, the
decoded pixel values are stored and used as context for predicting subsequent
tiles. This process repeats until all tiles have been processed.

\subsection{Spatial Context Prediction}
\label{subsec:context-prediction}

The spatial context predictor is a deep neural network that analyzes
incomplete $64 \times 64$ image patches and generates $32 \times 32$ images
that complete the original patch (see Figure~\ref{fig:kermit-tiles}). Our
architecture is based on the work of Pathak {\it et al.} who developed a
network that could inpaint missing tiles or random regions within a larger
patch~\cite{pathakCVPR16context}. Whereas their method was trained to
incorporate context from all directions, our network is trained exclusively to
predict the lower-right quadrant of an image patch to support raster order
encoding and decoding.

Figure~\ref{fig:context-predictor-arch} shows the architecture of our spatial
context predictor network. The 3-channel context patch is taken as input and
processed by four convolutional layers (stride = 2). Each of these layers
learns a feature map with a reduced resolution and a higher depth. A
``channel-wise, fully-connected'' layer (as described
in~\cite{pathakCVPR16context}) is implemented using a depthwise followed by a
pointwise convolutional layer. The goal of this part of the network is to
allow information to propagate across the entire tile without incurring the
full quadratic cost of a fully-connected layer. For our network, a
fully-connected layer would require 64 million parameters ($(4 \times 4 \times
512)^2$), whereas the channel-wise approach only requires 384 thousand ($4
\times 4 \times 8192 + 512 \times 512$), a 170x reduction. The final stage of
the network uses upsampled convolution (sometimes called ``deconvolution'',
``fractional convolution'', or ``up-convolution'') to incrementally increase
the spatial resolution until the last layer generates a 3-channel image from
the preceding $32 \times 32 \times 64$ feature map.

\subsection{Residual Encoding with Recurrent Networks}
\label{subsec:residual-encoding}

The context predictor typically generates accurate low-frequency data for each
new tile, but it is not able to recover many image details. To improve
reconstruction quality, the next step of our algorithm uses a second deep
network that learns to compress and reconstruct residual images. The
architecture of this network is based on recurrent auto-encoders and a binary
bottleneck layer (see Figure~\ref{fig:donald-rnn}). Specifically, we adopt the
``LSTM (Additive Reconstruction)'' architecture presented by Toderici {\it et
  al.}~\cite{Toderici2016}, except that where that paper trains the network to
compress full images, we train it to compress the residual within each tile
after running the context predictor.

\begin{figure}
  \centering
  \includegraphics[width=0.495\columnwidth]{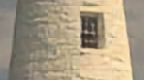}
  \hfill
  \includegraphics[width=0.495\columnwidth]{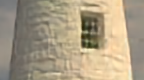}
  \caption{Block artifacts are visible when tiles are coded independently
    ({\it left}) but disappear when the spatial context predictor is used
    ({\it right}) [{\it Best viewed zoomed in].}}
  \label{fig:no-context}
\end{figure}

The encoder portion of the network uses one convolutional layer to extract
features from the input residual image followed by three convolutional LSTM
layers that reduce the spatial resolution (stride = 2) and generate feature
maps. Weights are shared across all iterations, and the recurrent connections
allow information to propagate from one iteration to the next. Our experiments
showed that the recurrent connections were vital and that this architecture
significantly outperformed a similar one made up of independent, non-recurrent
auto-encoders.

The binary bottleneck layer maps incoming features to $\{-1, 1\}$ using a $1
\times 1$ convolution followed by a tanh activation function. Following the
work of Raiko {\it et al.} on learning binary stochastic
layers~\cite{raiko:2015}, we sample from the output of the tanh ($P(b=1)=0.5
\cdot (1 + tanh(x))$) to encourage exploration in parameter space. When we
apply the trained network to images at run-time, however, we binarize
deterministically ($b=sign(tanh(x))$ with $b=1$ when $x=0$).

The decoder sub-network has the same structure as the encoder, except
upsampled convolution is used to increase the resolution of each feature map
by $2 \times$ in each layer. The final layer takes the output of the decoder
(a feature map with shape $32 \times 32 \times 64$) and uses a tanh activation
to map the features to three values in the range $[-1, 1]$. The output is then
scaled, clipped, and quantized to 8-bit RGB values ($R=round(min(max(R' \cdot
142 + 128, 0), 255))$). Note that we scale by 142 instead of 128 to allow the
network to more easily predict extreme pixel values without entering the range
of tanh with tiny gradients, which can lead to slow learning.

\subsection{Spatially Adaptive Bit Allocation}
\label{subsec:spatially-adaptive}

Adaptive bit allocation is difficult in existing neural network compression
architectures because the models are fully-convolutional. If such networks are
trained with all of the binary codes present, reconstruction with missing
codes can be arbitrarily bad. Our approach avoids this problem by sharing
information from the binary codes within each tile but not across tiles. This
strategy allows the algorithm to safely reduce the bit rate in one area
without degrading the quality of neighboring tiles.

\begin{figure}
  \centering
  \includegraphics[width=\columnwidth]{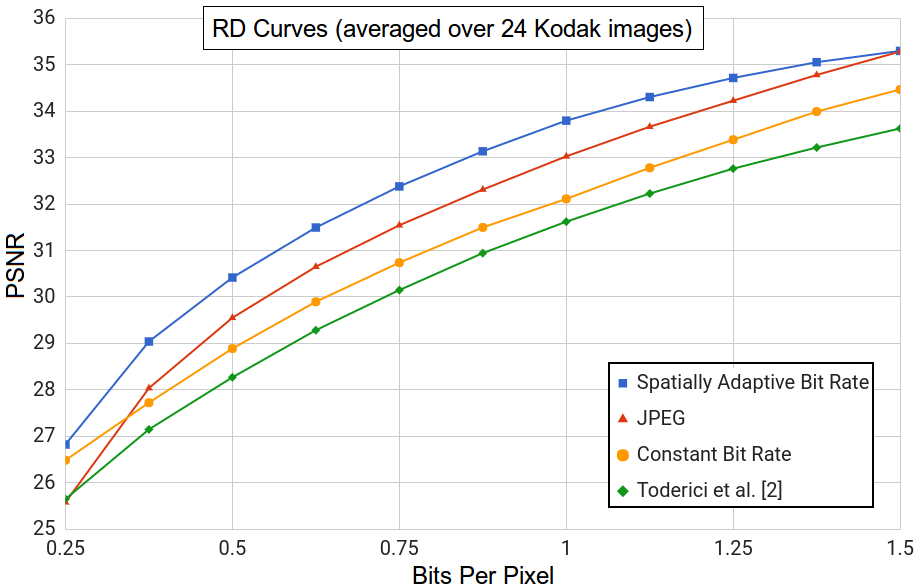}
  \caption{Using a constant bit rate, our approach shows a small PSNR
    improvement over the method in~\cite{Toderici2016} but only outperforms
    JPEG at very low bit rates. By adapting the bit rate to local image
    complexity, our method yields a higher mean PSNR across the full range
    (0.25 -- 1.5 bpp).}
  \label{fig:psnr-graph}
\end{figure}

One potential pitfall of a block-based codec is the possible emergence of
boundary artifacts between tiles. The spatial context predictor helps avoid
this problem by sharing information across tile boundaries without increasing
the bit rate (see Figure~\ref{fig:no-context}). In essence, the context
prediction network learns how to generate pixels that mesh well with their
context. Furthermore, since the predicted pixels are more detailed and
accurate near the context pixels, the network naturally acts to minimize
border artifacts.

\begin{figure*}
  \centering
  \includegraphics[width=\textwidth]{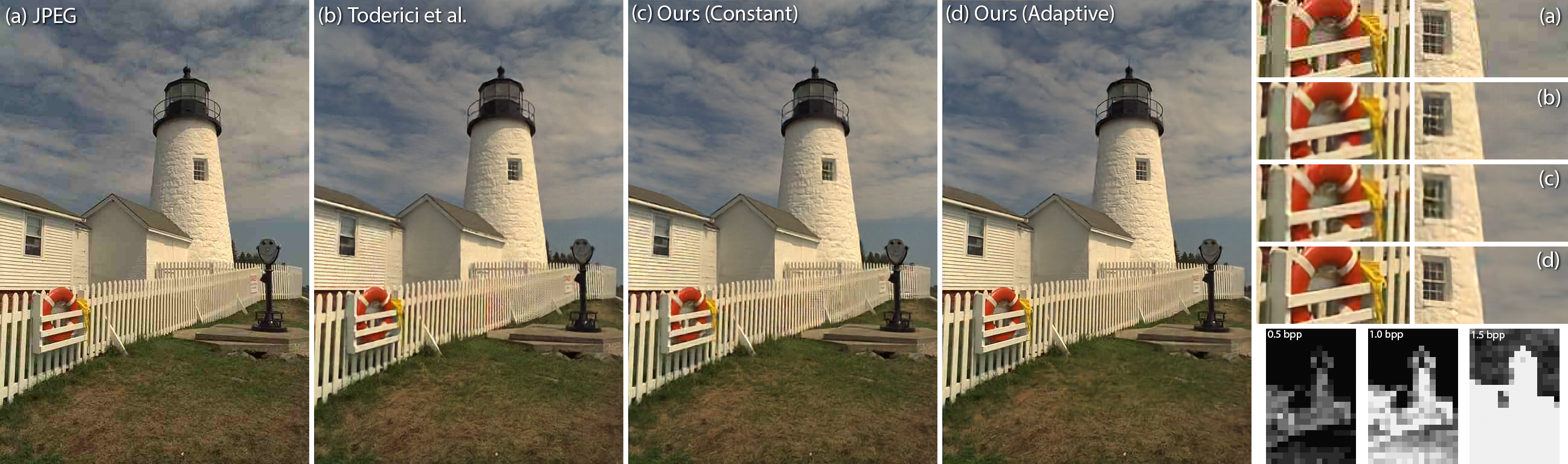}
  \caption{Reconstructions at 0.5 bpp: (a) JPEG (PSNR=29.552), (b) Toderici
    {\it et al.}~\cite{Toderici2016} (28.270), (c) our method with constant
    bit rate (28.890), and (d) our adaptive model (30.418). The far right
    (top) shows two zoomed-in regions for better comparison, while the bottom
    shows the adaptive bit rate mask calculated at three bit rates.}
  \label{fig:lighthouse}
\end{figure*}

Our approach for allocating bits across each image is straightforward. During
image encoding, each tile uses enough bits to exceed a specified target
quality level (compared to a target bit rate in the constant bit rate
case). The results presented below are based on a PSNR target, but any local
quality or saliency measure can be used (see the bottom-right of
Figure~\ref{fig:lighthouse} for examples of bit rate maps).

\subsection{Training and Run-Time Details}
\label{subsec:train}

Both the spatial context predictor and residual encoder networks were
implemented using Tensorflow~\cite{tensorflow} and trained using the Adam
optimizer~\cite{kingma:2014}. They are trained sequentially since the residual
encoder network learns to encode the specific pixel errors that remain after
context prediction. The training process used a mini-batch size of 32 and an
initial learning rate of 0.5 following an exponential decay schedule
($\beta=0.95$) with a step size of 20,000.

Our training data consists of $64 \times 64$ image patches cropped from a
collection of six million $1280 \times 720$ public images from the
web. Following the procedure described by Toderici {\it et
  al.}~\cite{Toderici2016}, we use the 100 patches from each image that were
most difficult to compress as measured by the PNG codec.

At run-time, the encoder process monitors the reconstruction error of each
tile and uses as few bits as possible to reach the target quality. This is
possible because the residual encoder is a recurrent network and can be
stopped after any step. Since each step generates additional bits, this
mechanism allows adapitve bit allocation and allows a single neural network to
generate encodings at different bit rates.

\section{Results and Evaluation}
\label{sec:evaluation}

We evaluated our approach with both quantitative and qualitative assessments
using the the Kodak image set~\cite{kodak}. Figure~\ref{fig:no-context}
includes two crops coded at 0.375 bits per pixel (bpp) that show the impact of
the spatial context predictor. Without it, each $32 \times 32$ tile is coded
independently and block artifacts are clearly visible.

The rate-distortion graph in Figure~\ref{fig:psnr-graph} shows PSNR values
averaged over the 24 images in the Kodak data set. The results show that our
approach outperforms the baseline neural network algorithm
from~\cite{Toderici2016} between 0.25 and 1.5 bpp. The spatially adaptive
version of our algorithm further increases reconstruction quality and
outperforms both of those models as well as JPEG~\cite{JPEG} across this bit
rate range.

Example images at 0.5 bpp are shown in Figure~\ref{fig:lighthouse}. JPEG shows
significant block artifacts and color shifts ({\it e.g.}, in the sky) not
present in the other images. Both Toderici {\it et al.} and our constant bit
rate reconstruction suffer from aliasing and a color shift on the fence, and
neither reconstructs the life buoy or yellow rope with much detail. Our
spatially adaptive method addresses all of these issues. Its reconstruction,
however, does have less detail in some visually simple but salient areas ({\it
  e.g.}, the mounted binoculars) and some neighboring regions have distracting
differences in the amount of retained detail ({\it e.g.}, where the fence
meets the grass). More sophisticated criteria for bit allocation that better
capture visual saliency will help in both cases and can be easily plugged in
to our algorithm.

Ten raters subjectively evaluated our results over the Kodak image set in a
pairwise study that included 24 images, four codecs, and six bit rates (0.25
-- 1.5 in 0.25 bpp increments) for a total of 8,640 image comparisons. In all
cases, the mean preferrence favored our adaptive algorithm over both the
constant bit rate version and the neural network baseline
from~\cite{Toderici2016}. Our adaptive algorithm was also preferred to JPEG at
0.25 and 0.5 bpp; elsewhere, the differences were not statistically
significant ($\alpha=0.05$).

\section{Conclusion and Future Work}
\label{sec:conclusion}

The primary goal of our current research is to combine deep neural networks
with spatial bit rate adaptation, which we think is vital for state of the art
compression results. By adopting a block-based approach, we are able to limit
the extent of local information sharing, which allows us to easily incorporate
a wide range of quality metrics to control local bit rate. Our experiments
show that explicit bit rate adaptation increases both quantiative and
subjective image quality assessments.

Our approach can be improved in many ways. Adaptively subdividing images
instead of using fixed $32 \times 32$ tiles will boost reconstruction quality
but requires more flexible network architectures. We can also adopt a
multiscale model where lower-resolution encodings act as a prior to guide the
predictions at higher resolutions. Better criteria for bit allocation should
yield significant quality improvements, particularly in terms of subjective
assessment. Finally, practical deployment will require additional research to
shrink the learned models and reduce their run-time requirements. Currently,
although the models produce higher quality compression results than JPEG,
their execution speed is much slower even when accelerated by modern GPU
hardware.

\balance

\bibliographystyle{IEEEbib}
\bibliography{biblio}

\begin{thebibliography}{10}

\bibitem{Jiang1999}
J.~Jiang,
\newblock ``Image compression with neural networks--a survey,''
\newblock {\em Signal Processing: Image Communication}, vol. 14, pp. 737--760,
  1999.

\bibitem{Toderici2016}
George Toderici, Damien Vincent, Nick Johnston, Sung~Jin Hwang, David Minnen,
  Joel Shor, and Michele Covell,
\newblock ``Full resolution image compression with recurrent neural networks,''
\newblock {\em CoRR}, vol. abs/1608.05148, 2016.

\bibitem{balle2016}
Johannes Ball\'{e}, Valero Laparra, and Eero~P. Simoncelli,
\newblock ``End-to-end optimization of nonlinear transform codes for perceptual
  quality,''
\newblock in {\em Picture Coding Symposium}, 2016.

\bibitem{toderici2015variable}
George Toderici, Sean~M O'Malley, Sung~Jin Hwang, Damien Vincent, David Minnen,
  Shumeet Baluja, Michele Covell, and Rahul Sukthankar,
\newblock ``Variable rate image compression with recurrent neural networks,''
\newblock {\em ICLR}, 2016.

\bibitem{gregor2016conceptual}
K.~{Gregor}, F.~{Besse}, D.~{Jimenez Rezende}, I.~{Danihelka}, and
  D.~{Wierstra},
\newblock ``{Towards Conceptual Compression},''
\newblock in {\em {NIPS}}, 2016.

\bibitem{jpeg2000}
``{Information technology--JPEG 2000 image coding system},''
\newblock Standard, International Organization for Standardization, Geneva, CH,
  Dec. 2000.

\bibitem{WebP}
Google,
\newblock ``{WebP}: Compression techniques
  ({http://developers.google.com/}\-{speed/}\-{webp/}\-{docs/}\-compression),''
  Accessed: 2017-01-30.

\bibitem{BPG}
F.~Bellard,
\newblock ``{BPG} image format ({http://bellard.org/}\-{bpg/}),'' Accessed:
  2017-01-30.

\bibitem{bull2014}
David~R. Bull, Ed.,
\newblock {\em Communicating Pictures: A Course in Image and Video Coding},
\newblock Academic Press, Oxford, 2014.

\bibitem{Long2014}
J.~Long, E.~Shelhamer, and T.~Darrell,
\newblock ``Fully convolutional networks for semantic segmentation,''
\newblock {\em CoRR}, vol. abs/1411.4038, 2014.

\bibitem{covell2017}
M.~Covell, N.~Johnston, D.~Minnen, S.J. Hwang, J.~Shor, S.~Singh, D.~Vincent,
  and G.~Toderici,
\newblock ``Target-quality image compression with recurrent, convolutional
  neural networks,''
\newblock {\em CoRR}, vol. abs/1705.06687, 2017.

\bibitem{pathakCVPR16context}
Deepak Pathak, Philipp Kr\"ahenb\"uhl, Jeff Donahue, Trevor Darrell, and Alexei
  Efros,
\newblock ``Context encoders: Feature learning by inpainting,''
\newblock in {\em CVPR}, 2016.

\bibitem{raiko:2015}
T.~Raiko, M.~Berglund, G.~Alain, and L.~Dinh,
\newblock ``Techniques for learning binary stochastic feedforward neural
  networks,''
\newblock {\em ICLR}, 2015.

\bibitem{tensorflow}
Mart\'{\i}n Abadi, Ashish Agarwal, Paul Barham, Eugene Brevdo, Zhifeng Chen,
  Craig Citro, Greg~S. Corrado, Andy Davis, Jeffrey Dean, Matthieu Devin,
  Sanjay Ghemawat, Ian Goodfellow, Andrew Harp, Geoffrey Irving, Michael Isard,
  Yangqing Jia, Rafal Jozefowicz, Lukasz Kaiser, Manjunath Kudlur, Josh
  Levenberg, Dan Man\'{e}, Rajat Monga, Sherry Moore, Derek Murray, Chris Olah,
  Mike Schuster, Jonathon Shlens, Benoit Steiner, Ilya Sutskever, Kunal Talwar,
  Paul Tucker, Vincent Vanhoucke, Vijay Vasudevan, Fernanda Vi\'{e}gas, Oriol
  Vinyals, Pete Warden, Martin Wattenberg, Martin Wicke, Yuan Yu, and Xiaoqiang
  Zheng,
\newblock ``{TensorFlow}: Large-scale machine learning on heterogeneous
  systems,'' 2015,
\newblock Software available from tensorflow.org.

\bibitem{kingma:2014}
D.~P. Kingma and J.~Ba,
\newblock ``Adam: {A} method for stochastic optimization,''
\newblock {\em CoRR}, vol. abs/1412.6980, 2014.

\bibitem{kodak}
Eastman Kodak,
\newblock ``Kodak lossless true color image suite ({PhotoCD PCD0992}),'' .

\bibitem{JPEG}
W.~Pennebaker and J.~Mitchell,
\newblock {\em {JPEG}: Still Image Compression Standard},
\newblock Kluwer Academic Publishers, 1992.

\end{thebibliography}

\end{document}